\documentclass[a4paper, conference]{IEEEtran}
\IEEEoverridecommandlockouts
\usepackage{cite}
\usepackage{amsmath,amssymb,amsfonts}
\usepackage{algorithmic}
\usepackage{graphicx}
\usepackage{textcomp}
\usepackage{xcolor}

\usepackage[left=1.57cm,right=1.57cm,top=0.95cm,bottom=2.54cm]{geometry}

\usepackage{hyperref}

\def\BibTeX{{\rm B\kern-.05em{\sc i\kern-.025em b}\kern-.08em
    T\kern-.1667em\lower.7ex\hbox{E}\kern-.125emX}}
\begin{document}

\title{
%Evolving, Fast and Slow for Feature Selection
Feature Selection with Evolving, Fast and Slow Using Two Parallel Genetic Algorithms 
}

\author{\IEEEauthorblockN{1\textsuperscript{st} Uzay Cetin}
\IEEEauthorblockA{\textit{Computer Engineering Department} \\
\textit{Istanbul Bilgi University}\\
Istanbul, Turkey \\
uzay00@gmail.com}
\and
\IEEEauthorblockN{2\textsuperscript{nd} Yunus Emre Gundogmus\IEEEauthorrefmark{1}\IEEEauthorrefmark{2}}
\IEEEauthorblockA{\IEEEauthorrefmark{1}
\textit{Tam Faktoring}}
\IEEEauthorblockA{\IEEEauthorrefmark{2}
\textit{Statistics Department} \\
\textit{Istanbul Marmara University }\\
Istanbul, Turkey \\
yemregun@gmail.com}
}

\maketitle

\begin{abstract}

Feature selection is one of the most challenging issues in machine learning, especially while working with high dimensional data. 
In this paper, we address the problem of feature selection and propose a new approach called \emph{Evolving Fast and Slow}. 
%
%The standard genetic algorithm is a building block for our new approach.
%
This new approach is based on using two parallel genetic algorithms having high and low mutation rates, respectively. 
\emph{Evolving Fast and Slow} requires a new parallel architecture combining an automatic system that evolves fast and an effortful system that evolves slow. 
With this architecture, exploration and exploitation can be done simultaneously and in unison.
\emph{Evolving fast}, with high mutation rate, can be useful to explore new unknown places in the search space with long jumps; and \emph{Evolving Slow}, with low mutation rate, can be useful to exploit previously known places in the search space with short movements.
Our experiments show that \emph{Evolving Fast and Slow} achieves very good results in terms of both accuracy and feature elimination.
\end{abstract}


\section{Introduction}
One of the most challenging issues in machine learning is to evaluate the importance of features in high dimensional data~\cite{survey, survey2}. 
Usually, data scientists gather with domain experts and make use of their business knowledge 
in order to distinguish significant features from non-significant features.
When that is not enough or not even possible, traditional feature selection algorithms can be used. 
Those algorithms are in general based on statistics such as chi-square test~\cite{chi2}
or based on decision tree approaches such as extra-tree classifier~\cite{extratree} or based on dimension reduction techniques such as principal component analysis~\cite{pca}.
In this paper we address the same problem of feature selection and propose a new approach based on genetic algorithms.

\subsection{The Use of Genetic Algorithms}
Genetic algorithm is developed by John Holland~\cite{holland}. to understand and imitate the adaptive processes of natural systems.
Genetic algorithms can be used for various optimization problems. 
In this paper, we adapt it for a well-known challenging machine learning problem, referred to as 
\emph{feature selection}. Feature selection problem is nothing but finding the most significant subset of features in a given dataset.
Suppose a dataset has 300 features, is it possible to do an effective search for the best subset of features using traditional \emph{for} loops? The answer is no.
Because, the number of possible combinations for 300 features is $2^{300} = (2^{10})^{30} > (10^{3})^{30} = 10^{90} $. That is far greater than $10^{82} $ which equals to the estimated number of atoms in the known, observable universe~\cite{atom}. As Salvatore Mangano put it~\cite{Chaturvedi},
\begin{quote}
``Genetic Algorithms are good at taking large, potentially huge search spaces and navigating them, looking for optimal combinations of things, solutions you might not otherwise find in a lifetime."
\end{quote}
That is why we need heuristic algorithms, for feature selection, such as genetic algorithms.

\begin{figure}[!h]
 \centering
  \includegraphics[width=0.95\linewidth]{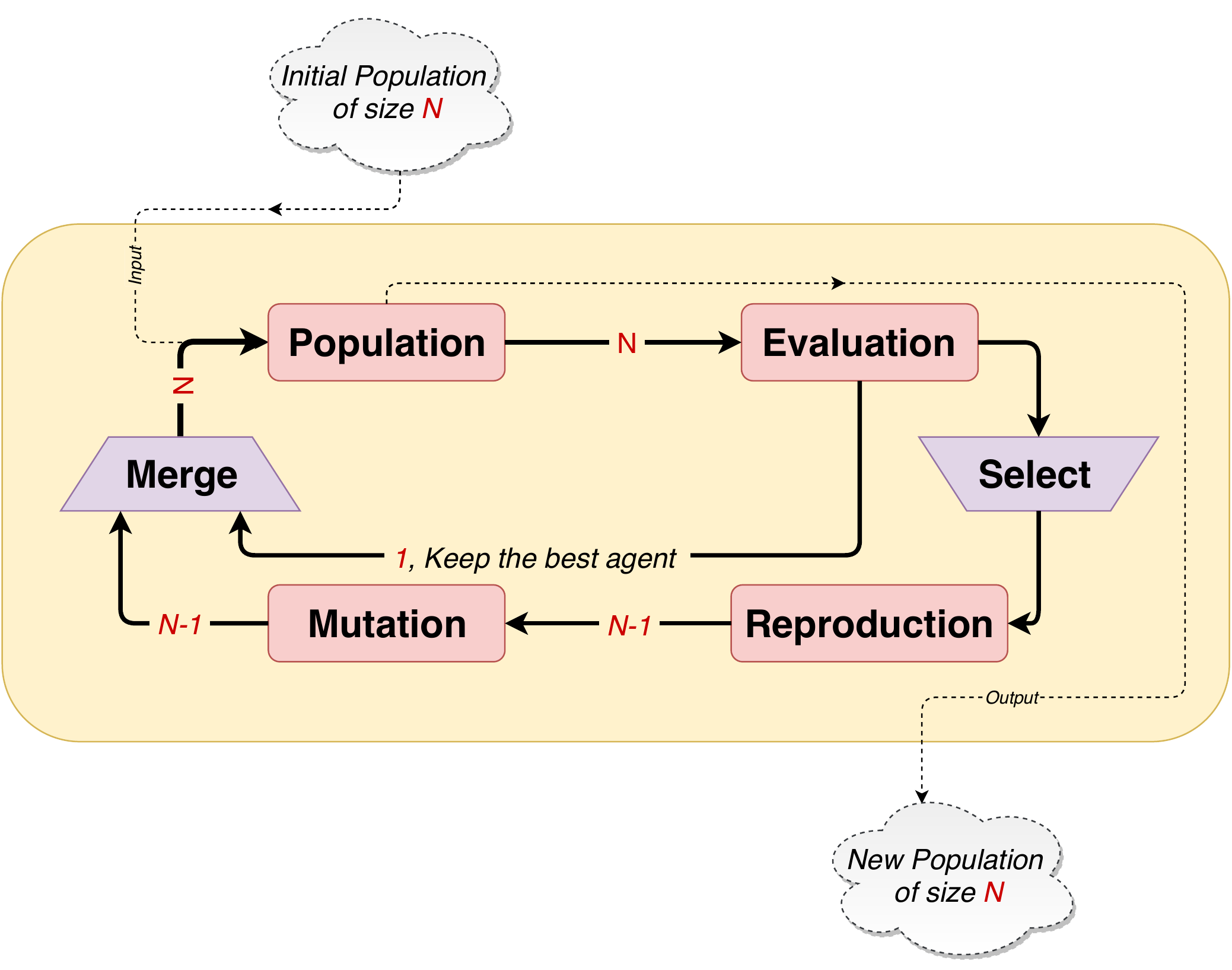}
  \caption{Components of Genetic Algorithm}
  \label{fig:Components}
\end{figure}
\subsection{Components of Genetic Algorithms}
The first step before starting using genetic algorithm is to determine
how to properly encode the candidate solutions as chromosomes. 
Then the process starts with the random creation of the initial population which is composed of $N$ chromosomes
as shown in Figure.~\ref{fig:Components}.
The rest of the algorithm is composed of the following steps applied in a loop, 

\begin{enumerate}
\item
Evaluation step for quantifying the fitness of each candidate chromosome.
\item
Selection of parents for reproduction based on their fitness values.
\item
Usage of genetic operators (mutation, recombination)  to create $N-1$ new offspring which are slightly modified versions of the mixture of their parents' chromsomes.
\item
Merging $N-1$ new offspring with the previous best chromosome, in order not to decrease the ovarall performance due to the inherent randomness in the process.
\item 
if any termination criteria is not met, going back to step 1 with the generated new population, whose size is $N$ again.
\end{enumerate}
\section{Genetic Algorithm For Feature Selection}
\label{sec:GA}
%\subsection{The Use of Genetic Algorithms}

\begin{figure}[!h]
 \centering
  \includegraphics[width=0.95\linewidth]{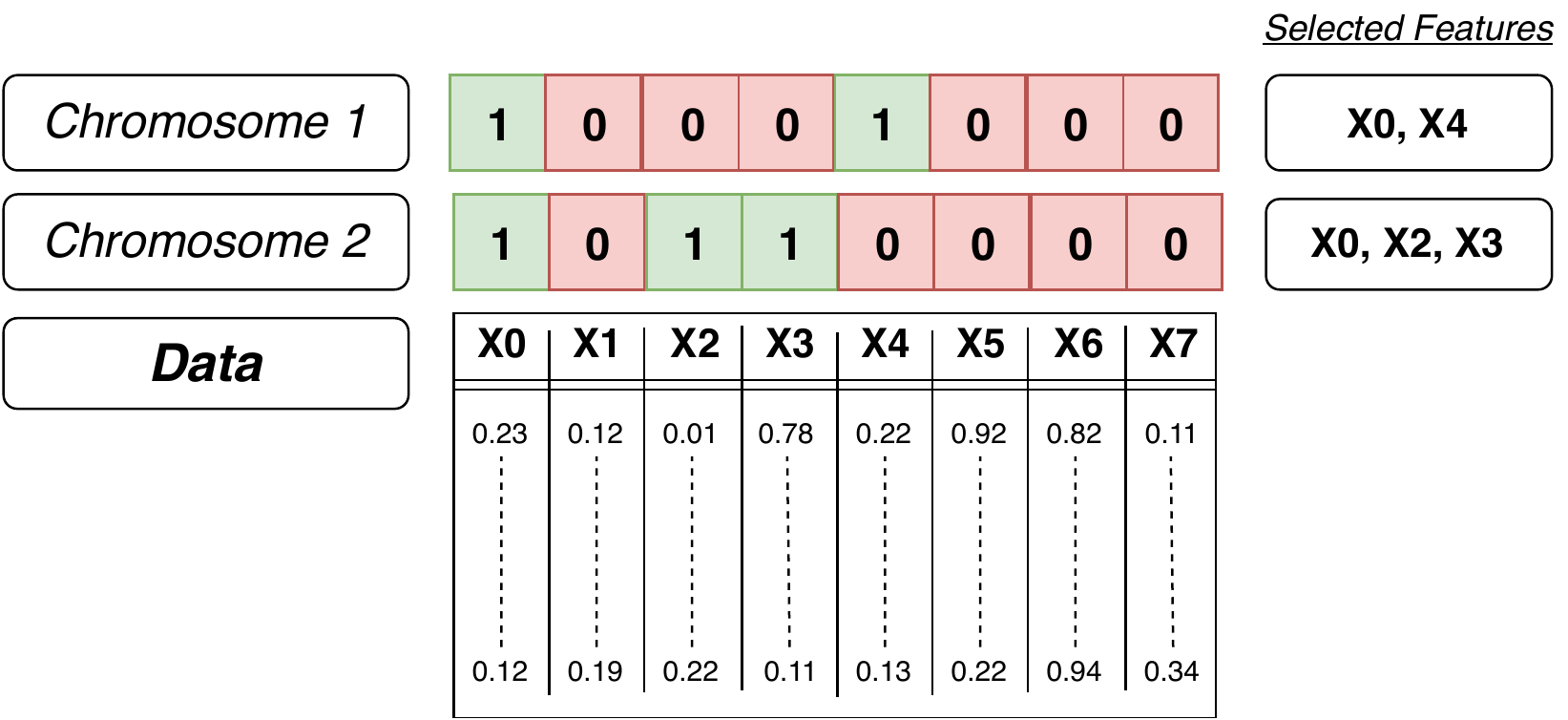}
  \caption{Binary encoding for the chromosomes where 1 represents selected feature and 0 represents disregarded feature. In this example, \emph{chromosome 1} has value of one at indices 0 and 4 which means $0^{th}$ and $4^{th}$ features are selected. In a similar manner,  
  \emph{chromosome 2}  selects $0^{th}$, $2^{nd}$ and $3^{rd}$ features.
}
  \label{fig:chromosome}
\end{figure}

\subsection{Encoding}
Genetic algorithm for feature selection also starts with the correct encoding.
The chromosome should contain information about which features are selected and which ones are omitted.
That can be done with binary encoding where
genes can have either a value 1 (for selected feature) or 0 (for omitted feature). 
As it can be seen in Figure.~\ref{fig:chromosome}, value 1 represents that the corresponding feature is selected and value 0 means that it is disregarded.

\subsection{Evaluation and Fitness}

Population is composed of several different instances of chromosomes. The goal of feature selection with genetic algorithm, is to find out the best chromosome (which corresponds to the best subset of features) that maximizes the cross validation accuracy score with the minimum number of features.
That requires two simultaneous optimization processes at the same time.
That is why we define the fitness function as follows:
\begin{equation}
fitness(x) = \alpha \times score(x)+ (1 - \alpha) \times (1 -  \frac{N_{x}}{N_{all}}) 
 \label{eq:fitness}
\end{equation}
For a given specific chromosome $x$, $N_x$ represents the number of selected features where 
$N_{all}$ is the number of all features within the given dataset. (In Figure.~\ref{fig:chromosome} for $x = chromosome \; 1$, $N_x = 2$ and $N_{all} = 8$.)

In Equation~\ref{eq:fitness}, $score(x)$ is the cross validation accuracy score computed with the selected $N_x$ features. 
We introduce a hyper-parameter called  $\alpha$ which determines the importance given to the accuracy score. Thus $(1 - \alpha)$ determines the importance given to the
feature elimination.
One advantage of this fitness score is that it is in the range [0,1], since both $score(x)$ and  $1 -  \frac{N_{x}}{N_{all}}$ is in between 0 and 1.
if $\alpha = 1$, $fitness(x)$ becomes equal to the cross validation accuracy score, $score(x)$.
Note also that, while the selected number of features decreases, that is $N_x$ approaching to zero, fitness increases.  

\begin{table}[htp]
\begin{center}
\begin{tabular}{|c|c|c||c|}
\hline
chromosome & score(x) & $N_x$ & fitness(x) \\
\hline
\hline
$x_1$ & 0.80 & 400 & 0.70 \\
\hline
$x_2$ & 0.80 & 1000 & 0.40 \\
\hline
$x_3$ & 0.82 & 1000 & 0.41 \\
\hline
\end{tabular}
\end{center}
\caption{For $\alpha = 0.5$ and $N_{all} = 1000$}
\label{table:fitness}
\end{table}%
In Table.~\ref{table:fitness}, we show the situation where $\alpha = 0.5$ for a fictitious dataset. $\alpha = 0.5$ means that feature elimination is as crucial as accuracy score. In that case, even if chromosome $x_3$, has higher accuracy score than chromosome $x_1$, its fitness is lower. That is because chromosome $x_3$ can not reduce the number of features at all.
With this hyper-parameter $\alpha$, our algorithm gains the ability to automatically find a balance between proper number of features and accuracy. Most of the existing feature selection algorithms requires the number of features given as an input \cite{survey}. But distinctively our method can decide the number of features on its own.

\subsection{Selection of Parents and Reproduction}
\begin{figure}[!h]
 \centering
  \includegraphics[width=0.6\linewidth]{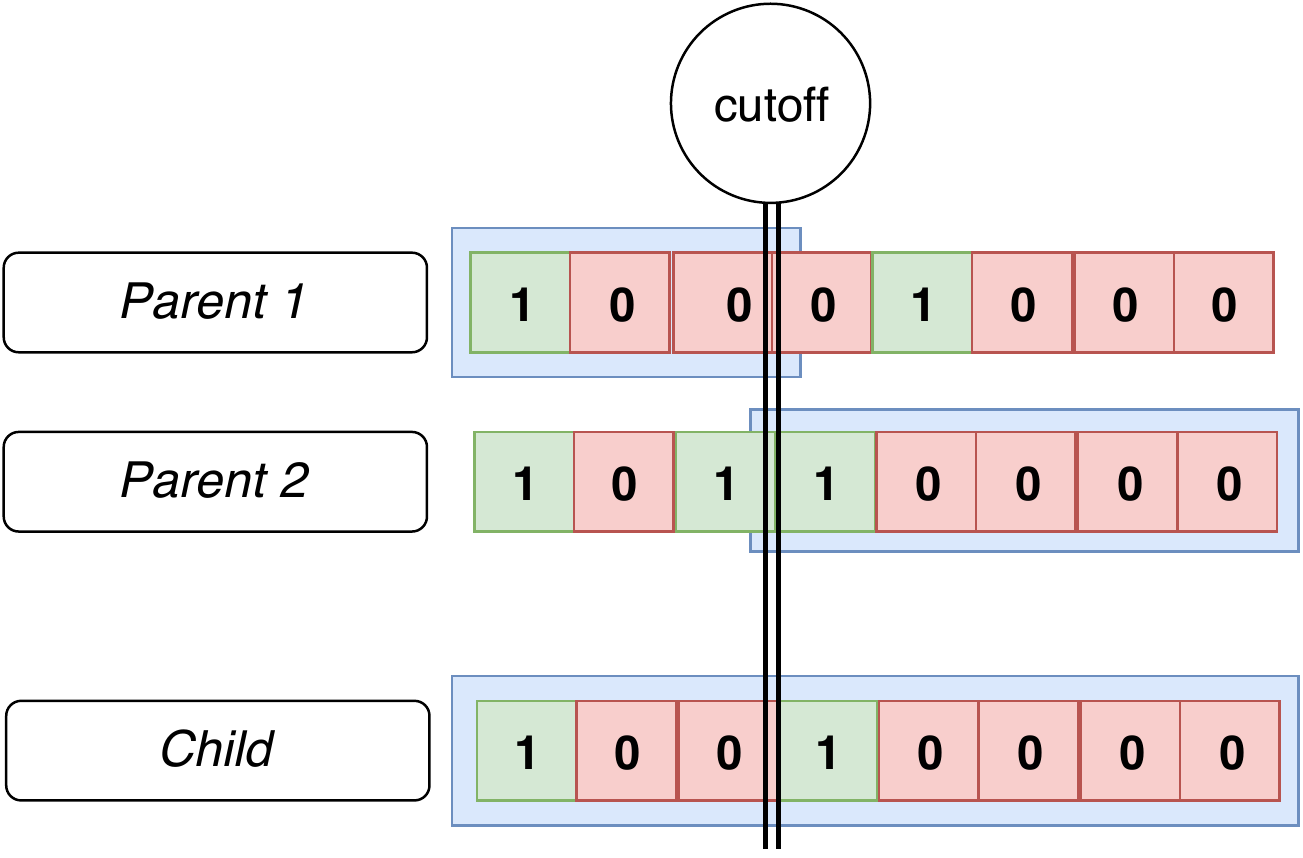}
  \caption{Child gets the first part of the first parent's chromosome  (from beginning to the cutoff point) and gets the second part of the second parent's chromosome  (from the cutoff point to the end) during recombination.}
  \label{fig:child}
\end{figure}

To generate the new population,
two consecutive genetic operators are applied: \emph{recombination} and \emph{mutation}.
$N-1$ different pairs of parent chromosomes are selected based on their fitness value to create $N-1$ offsprings.
Selection is done according to the following rule: 
higher the fitness value, higher the chance to get selected as a parent.
Once two parents are selected, a random cutoff point is determined for recombination of parent's chromosomes as shown in Figure.~\ref{fig:child}.
Recombination is followed by mutation.
With some small probability, specified by the mutation rate $\mu$, gene values are mutated.
Mutation occurs with flipping one to zero for previously selected features or
flipping zero to one for previously unselected features.
While generating new population, we keep the best in order not to decrease performance.
As a result,
new population is composed of $N-1$ new \emph{mutant} offsprings and $1$ \emph{best} parent 
chromosome which had the highest fitness within the previous population. 
This process ensures that new population does not get worse in terms of performance.
Overall architecture for genetic algorithm can be seen in Figure.~\ref{fig:Components}.

\section{Evolving, Fast and Slow}
We propose a new feature selection algorithm design with genetic algorithms, inspired from the international bestseller book \emph{Thinking, Fast and Slow}~\cite{FastSlow} written by
Nobel Laureate Daniel Kahneman. He is  one of the
world's most influential living psychologist~\cite{theguardian}.

Kahneman introduced the metaphor of of two agents, called System1 and System 2, which respectively produce fast and slow thinking. 
System 1 (Thinking Fast) can be considered as an automatic, intuitive, “gut reaction” way of thinking 
whereas System 2 (Thinking Slow) can be considered as a lazy controller with focus and critical thinking.
According to him,
\begin{quote}
``System 1 (Thinking Fast) continuously generates suggestions for System 2 (Thinking Slow): impressions, intuitions, intentions, and feelings. If endorsed by System 2, impressions and intuitions turn into beliefs, and impulses turn into voluntary actions."
\end{quote}
In this paper, we adapt these two fictitious characters and design a parallel architecture of two genetic algorithms \emph{evolving fast and slow}.

\begin{figure}[!h]
 \centering
  \includegraphics[width=0.95\linewidth]{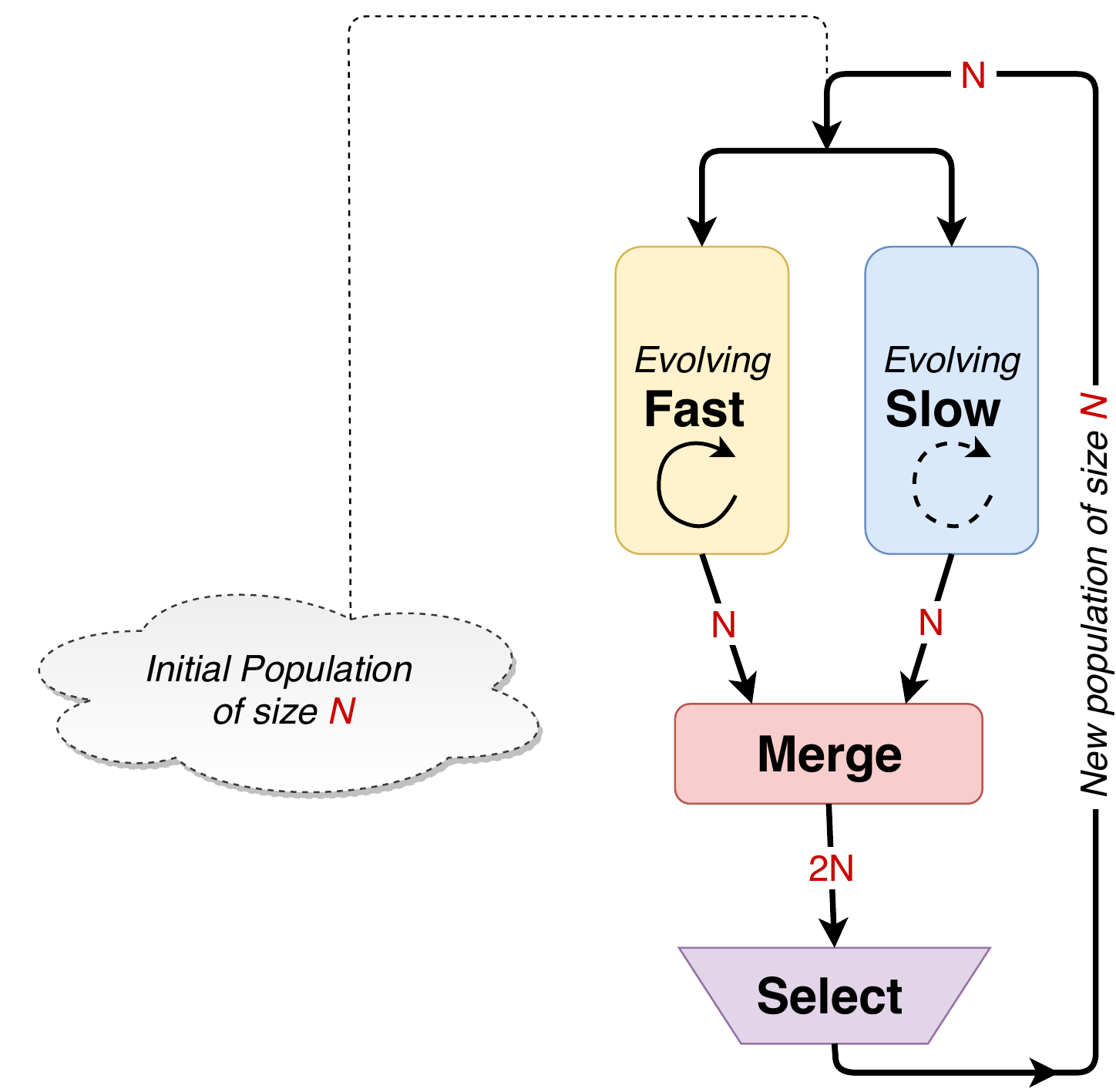}
  \caption{Evolving, fast and slow 
  with two parallel genetic algorithms 
  having high and low mutation rates, respectively. 
}
  \label{fig:FastSlow}
\end{figure}

The standard genetic algorithm described in Section.~\ref{sec:GA}, is the building block for our new \emph{Evolving Fast and Slow} architecture. This architecture can be seen in detail, in Figure.~\ref{fig:FastSlow}.
It is composed of two parallel standard genetic algorithms (GAs), taking the same population of $N$ chromosomes as input.
\begin{itemize}
\item 
Those two GAs, differ only in their mutation rate. We call \emph{Fast}, the GA that has high mutation rate and the other as \emph{Slow} who has low mutation rate.

\item
\emph{Fast} and \emph{Slow} GAs, evolve separately in a distributed manner for a given number of generations, $G_{in}$. 

\item
They produce completely different new offspring populations. The critical step is to merge these two  populations into one population of size $N$. 

\item 
Selection is done by taking the first $N$ most successful chromosomes among the merged $2N$ chromosomes.

\item 
 if any termination criteria is not met, going back to step 1 with the generated
new population composed of the $N$ most successful chromosomes. 
\end{itemize}
\subsection{Exploration and Exploitation in Unison}
This new parallel architecture is a combination of an automatic system that \emph{evolves fast} and an effortful system that \emph{evolves slow}.
This architecture solves the exploration and exploitation dilemma, by taking the advantage of \emph{evolving fast and slow}.
 \emph{Evolving fast}, with high mutation rate, can be useful to explore new unknown places in the search space with long jumps; and  \emph{Evolving Slow},  with low mutation rate, can be useful to exploit previously known places in the search space with short movements.

\subsection{Efficiency}
Computational complexity does not change while evolving fast and slow. 
The computational complexity of using one genetic algorithm with a population that is composed of $2N$ chromosomes, is exactly the same as running two genetic algorithms with population size $N$. 
Moreover, two genetic algorithms can be run in parallel, which decreases the running time to half.

\section{Results}

We create a toy dataset for binary classification task as a controlled experiment, to better understand the effect of different mutation rates.
Dataset is generated such that it has 10000 samples and 50 Features.
Target value $y$ is 1, if the mean of first 10 significant features
is greater than $0.5$ and 0 otherwise.
So, only the first 10 feature is significant to determine the target value $y$ and the rest 40 features are nothing but noise.
An efficient feature selection algorithm, should be capable of finding these significant features.

\begin{figure}[!h]
 \centering
  \includegraphics[width=\linewidth]{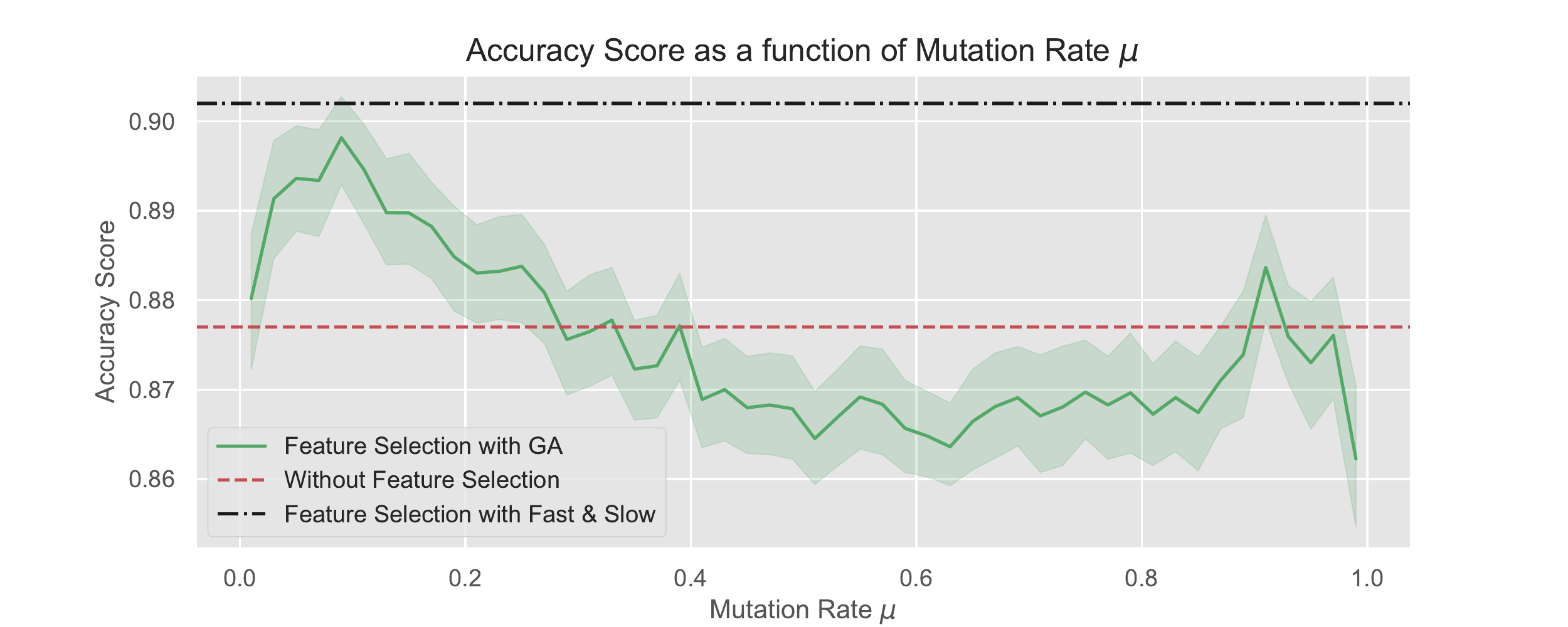}
  \caption{Average accuracy score on the generated toy dataset.}
  \label{fig:GAs}
\end{figure}

Experiments are done for consecutive fifty $\mu$ values from $0.01$ to $0.99$ with incremental steps of $0.02$.
For these experiments, we always used scikit-learn's
logistic regression implementation as the estimator~\cite{scikit-learn}.
We also fixed $\alpha$, importance given to the accuracy score, to 1 during these experiments. 
For each mutation rate $\mu$ value, we run genetic algorithm for feature selection 
during 20 generations, where the population size is also 20.
Average of fifty runs, can be seen in Figure.~\ref{fig:GAs} with the straight green line.
The maximum accuracy score is $0.898$ attained with $\mu = 0.09$.
Since this is not a complicated dataset, the accuracy score without feature selection is $0.877$.
For these dataset, using a higher mutation rate than $0.3$ decreases the accuracy score, even below the base accuracy score attained without any feature selection (shown as dashed red line in Figure.~\ref{fig:GAs} ).
Very interestingly, accuracy score does not continuously drop while mutation rate increases.
After $0.85$, accuracy score again increases above the base accuracy score.
It was a little bit unexpected at first.
But this is due to the process of keeping best chromosome from generations to generations.
And, while mutation rate is very high,
genetic algorithm can search the space with very long jumps without loosing the best candidate solution found so far.

Finally, \emph{Evolving Fast and slow}  with mutation rates $\mu_{Slow} = 0.1$ and $\mu_{Fast} = 1$
approach achieves significantly better performance in accuracy rate ($0.902$) on the average. It is indicated with dash-dotted line in black color in the Figure.~\ref{fig:GAs}.
We also looked at the number of features.
In this experiment with \emph{Evolving Fast and slow}, $\alpha $ was $1$ too. 
So the process did not care much about the reduction of features
and kept $27.8$ features on the average.
Detailed analysis show that it always finds the significant features and get rid of half of the noisy features with $\alpha = 1$. 
\begin{figure}[!h]
 \centering
  \includegraphics[width=\linewidth]{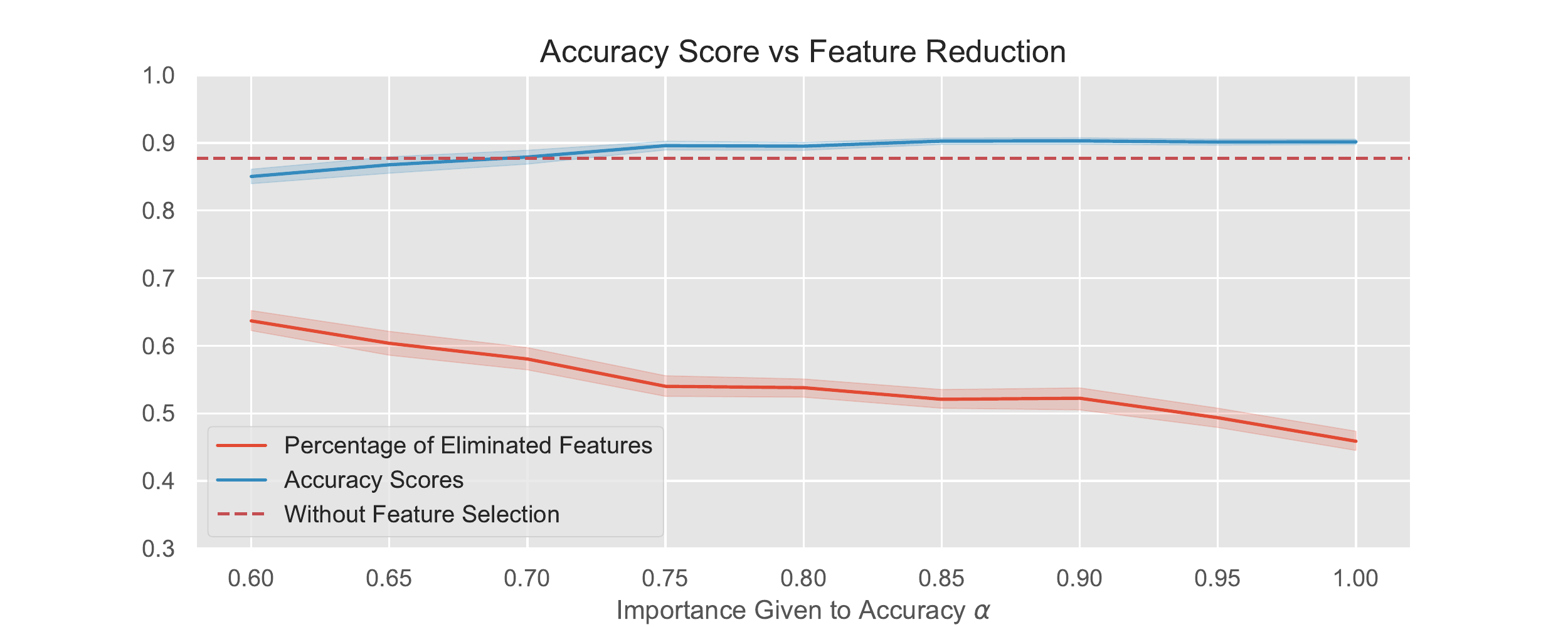}
  \caption{Average accuracy score on the generated toy dataset.}
  \label{fig:alpha}
\end{figure}

We did further experiments with
\emph{Evolving Fast and slow} with mutation rates $\mu_{Slow} = 0.1$ and $\mu_{Fast} = 1$,
to understand the effect of $\alpha$, the importance given to the accuracy score. 
Figure.~\ref{fig:alpha} shows that generally greater the $\alpha$, greater the average accuracy score.
But the maximum accuracy score is attained with $\alpha = 0.90$, which is $0.903$.
This average score is calculated over 50 runs.
We see in Figure.~\ref{fig:alpha} that 
more and more features are eliminated 
with smaller $\alpha$ value.
That is more and more noisy features are eliminated.
This is a natural consequence of the fitness equation.
While $\alpha$ gets smaller, importance for feature elimination increases in Equation~\ref{eq:fitness}.

All these hyper-parameters of $\alpha$, $\mu_{Slow}$ and $\mu_{Fast}$ requires to be searched for different datasets. In this paper, we showed that $\alpha = 0.90$ works best for $\mu_{Slow} = 0.10$ and $\mu_{Fast} =1$. 
We have chosen $\mu_{Slow} = 0.10$ and $\mu_{Fast} =1$ because they have shown good results individually on feature selection with standard genetic algorithm. 
We leave finding the best parameters for $\mu_{Slow}$ and $\mu_{Fast}$ for  different real datasets with \emph{Evolving Fast and slow} as a future work. 
But we show our results for 2 different real datasets in the next section.

\subsection{Turkish Political Dataset}
Turkish political climate dataset~\cite{siyasalyonelim} contains 
information about the opinion of the individual voters on a variety of political issues related to Turkish politics. 
Dataset has 885 rows and 14 columns.  
More information about the dataset, can be found in article~\cite{siyasalyonelim}.

\begin{table}[h]
\centering
\caption{Political Dataset with 2 Parties, AKP and CHP}
\begin{tabular}{|c||c|c|}
\hline
\textbf{Political Dataset }               & Accuracy           & Number of Features \\
\hline
\hline
Fast \& Slow            & $80.93 \pm 0.061$  & $1.55 \pm 0.739$  \\
\hline
\hline
GA $\mu = 0.10$ & $78.14\pm 0.072$  & $1.4 \pm 0.663$  \\
\hline
GA $\mu = 0.90$  & $80.62 \pm 0,064$ & $2.3 \pm 0.842$   \\
\hline
Without Feature Selection  & $85.95 \pm 0.005$ & 14  \\
\hline
\end{tabular}
\label{table:Political2}
\end{table}

\begin{table}[h]
\centering
\caption{Political Dataset with 6 parties}
\begin{tabular}{|c||c|c|}
\hline
\textbf{Political Dataset }               & Accuracy           & Number of Features \\
\hline
\hline
Fast \& Slow            & $38.55 \pm 0.060$ & $1.7 \pm 0.714$  \\
\hline
\hline
GA $\mu = 0.10$ & $37.53 \pm 0.049$ & $1.1 \pm 0.300$  \\
\hline
GA $\mu = 0.90$  & $36.97 \pm 0.057$ & $1.7 \pm 0.714$  \\
\hline
Without Feature Selection  & $43.67 \pm 0.013$ & 14  \\
\hline
\end{tabular}
\label{table:Political6}
\end{table}
Results are computed over 20 realizations for $\alpha = 0.9$ with Random Forest Classifier~\cite{scikit-learn} in Tables~\ref{table:Political2} and~\ref{table:Political6}.
The second row of the table shows the results for  \emph{Evolving Fast and Slow}
where  $\mu_{Slow}=0.10$ and $\mu_{Fast} = 0.90$.
The third and fourth rows show the results for standard genetic algorithms, for mutation rates $\mu=0.10$ and $\mu = 0.90$ respectively. 
And the last row shows the results without any feature selection.
All results are given with mean and standard deviation values over 20 realizations.

Our experiments show that \emph{Evolving Fast and Slow} achieves better results in terms of both accuracy and feature elimination in these datasets.
%Average of 20 realizations for $\alpha = 0.9$ on

\subsection{Tam Faktoring Cheque Dataset}
Tam Faktoring Cheque Dataset has 50 different columns~\cite{Tam}. Each column represent different features, 
\begin{table}[h]
\centering
\caption{Financial Dataset}
\begin{tabular}{|c||c|c|}
\hline
\textbf{Financial Dataset }               & Accuracy           & Number of Features \\
\hline
\hline
Fast \& Slow           & $67.73 \pm 0.029$ & $14.7\pm 1.873$   \\
\hline
\hline
GA $\mu = 0.10$ & $67.46 \pm 0.023$ & $13.65 \pm 2.006$  \\
\hline
GA $\mu = 0.90$   & $67.89 \pm 0.029$ & $16.0 \pm 2.0$  \\
\hline
Without Feature Selection  & $68.83 \pm 0.006$ & 50  \\
\hline
\end{tabular}
\label{table:Financial}
\end{table}
some of which are \emph{cheque value, due date,
customer's previous credit information, credit application count, all open credits balance, KKB credit score, open credit cards debt, etc. (all the customer data masked in accordance with KVKK(Personal Data Protection Law)).
}

Results are again computed over 20 realizations for $\alpha = 0.9$ for the financial dataset~\cite{Tam} with Random Forest Classifier~\cite{scikit-learn} in Table~\ref{table:Financial}.
In this particular dataset, standard genetic algorithm with mutation rate $\mu = 0.90$ is slightly better in terms of accuracy but by using more features.
Our experiments show that \emph{Evolving Fast and Slow} achieves very good results in terms of both accuracy and feature elimination.

\section{Conclusion}
Feature selection and algorithm selection are the most required skills, expected from data science experts.
We have proposed and implemented a new feature selection algorithm design called \emph{Evolving Fast and Slow}
for classification problems in machine learning.
The standard genetic algorithm is a building block for our new approach.
This new approach is based on using two parallel genetic algorithms having high and low mutation rates, respectively.
Our approach can easily be extended to regression problems.
%
%We leave that for a future work.
%
%Our method is based on genetic algorithms.
%
There are several benefits of genetic algorithms, first of all it is 
inherently parallel and can be easily distributed.
It always finds a solution.
With correct hyper-parameter settings, solutions found gets better.
We have shown the best hyper-parameter settings for a toy dataset and applied them to two different real datasets~\cite{siyasalyonelim, Tam}.
We leave a systematic search of hyper-parameters for different kind of real datasets as a future work.
One of the key advantage of this work, is that this method can decide the number of features to be selected on its own, whereas most of the existing feature selection algorithms requires the number of selected features as an input.
Apart from feature selection, algorithm selection can also be done via genetic algorithms.
That is another interesting research question that can be addressed with our proposed method.

%We leave algorithm selection also for a future work.

%

\section*{Acknowledgment}
This study is accomplished as a term project of "Free Artificial Intelligence Course to Young People" in Sariyer Akademi, given by Uzay Cetin~\cite{egitim}. 
We thank Sariyer Akademi and Sami Gorey for their support~\cite{sariyer}.

\end{document}